\documentclass[11pt]{tibop-article}

\TIBauthor[D'Souza \& Ahmed]{%
    Jennifer D'Souza\affOne\orcidlink{0000-0002-6616-9509}\and
    Fahad Ahmed\affOne\orcidlink{0000-0001-9201-1580} \and
    Cecilia Andrea Bustamante Andrade\affTwo\orcidlink{0009-0006-2103-5941} \and
    Lina Frolova\affThree\orcidlink{0009-0004-7381-6892} \and
    Poorani Gnanasambandan\affTwo\orcidlink{0000-0001-9850-7966} \and
    Dilshad Hussain\affFour \and
    Muhammad Uzair Khan\affFive\orcidlink{0009-0002-1004-7526} \and
    Nkembeng Kevin Nkengfoa\affSix\orcidlink{0009-0001-0398-2631} \and
    Paul Praveen J.\affSeven\orcidlink{0000-0002-7149-8833} \and
    Fabio Priante\affEight\orcidlink{0000-0001-7052-8570} \and
    Sjoerd Franciscus van der Werf\affTwo\orcidlink{0009-0007-2496-3193} \lastand
    Thomas Frederik Jan van Roeden\orcidlink{0009-0007-9682-4723}
}

\TIBaffiliations{%
    \affOne Data Science and Digital Libraries, 
    TIB Leibniz Information Centre for Science and Technology, 
    Hannover, Germany\medskip\\
    \affTwo Department of Applied Physics and Science Education, Eindhoven University of Technology, Eindhoven, Netherlands\medskip\\
    \affThree Department of Biology, Chemistry, and Pharmacy, Freie Universität Berlin, Berlin, Germany\medskip\\
    \affFour HEJ Research Institute of Chemistry, International Center for Chemical and Biological Sciences, University of Karachi, Karachi, Pakistan\medskip\\
    \affFive School of Interdisciplinary Engineering and Sciences, National University of Sciences and Technology, Islamabad, Pakistan\medskip\\
    \affSix Department of Chemistry, University of Warwick, Coventry, United Kingdom\medskip\\
    \affSeven Department of Physics, PSG College of Technology, Coimbatore, Tamil Nadu, India\medskip\\
    \affEight Department of Chemistry and Materials Science, Aalto University, Helsinki, Finland\medskip\\
    *Correspondence: jennifer.dsouza@tib.eu
}

\TIBtitle{A Pathway to General-Purpose Scientific AI: Multimodal Comprehension of Scientific Images}

\TIBbundlename[ConfAbbrev]{Full conference name}

\TIBbundlename[
Open Conf Proc X (2025)
``ICDAR 2026 Competition on Information Extraction from Atomic Layer Deposition/Etching (ALD/E) Scientific Figures''
]{%
ICDAR 2026 Competition on Information Extraction from Atomic Layer Deposition/Etching (ALD/E) Scientific Figures
}

\TIBconferencesession{Sci-ImageMiner 2026 Competition Proceedings}

\TIBsubmitteddate{2026-08-10}

\TIBabstract{%
Scientific figures and tables encode essential experimental evidence, yet remain difficult for digital libraries and multimodal AI systems to retrieve and interpret. The ALD/E-ImageMiner benchmark and ICDAR 2026 Competition on Information Extraction from Atomic Layer Deposition/Etching Scientific Figures provide 1,951 figures from 205 publications, expert-annotated for classification, data table extraction, summarization, and visual question answering. In these companion proceedings, we present a forward-looking perspective on how the benchmark can guide future scientific-image challenges. We examine how its tasks probe capabilities from visual and quantitative reading to domain-grounded reasoning and evidential justification, and how Bloom-informed question design can support deeper scientific understanding. We propose ``scientific conceptual understanding from images'' as a long-term benchmark objective, with future directions including broader domains and figure types, contextual and cross-document synthesis, hypothesis evaluation, provenance, uncertainty, counterfactual grounding, and open-ended multimodal research. This perspective connects the ICDAR 2026 challenge to a broader agenda for machine-actionable scientific visual knowledge and verifiable multimodal scientific AI.
}

\TIBkeywords{scientific images, multimodal AI, vision-language models, digital libraries, benchmarks, scientific reasoning}

\usepackage{xfrac}
\usepackage{comment}
\usepackage{amsmath}

\begin{document}

\maketitle

\begin{itemize}[]
\item \textbf{URL}: \url{https://sciknoworg.github.io/ALD-E-ImageMiner/}
\item \textbf{Huggingface}:  \url{https://huggingface.co/datasets/SciKnowOrg/ALD-E-ImageMiner}
\item \textbf{License}: Mixed right non-commercial
\end{itemize}

\section{Introduction}

Scientific communication is intrinsically multimodal. Researchers convey experimental observations, quantitative results, structural relationships, and methodological details through charts, spectra, microscopy images, molecular and reaction diagrams, process flows, and apparatus schematics. These visual artifacts are not merely illustrations of the surrounding text; they often constitute the primary record of the evidence on which scientific claims are based. For digital libraries, this creates a fundamental challenge: scholarly knowledge cannot be comprehensively indexed, retrieved, or analyzed when figures and tables remain accessible only as undifferentiated image objects. Future scientific AI systems must therefore be able to extract and reason over information represented jointly in visual and textual forms, from interpreting spectroscopic plots and microstructural features to understanding experimental configurations \cite{alampara2025probing}. Without this capability, both AI-assisted scientific analysis and the transformation of digital libraries into multimodal knowledge infrastructures remain incomplete.

Recent vision--language models have broadened natural-language access to visual content, but their performance on scientific figures remains uneven \cite{alampara2025probing,ahmed2026icdar}. Scientific interpretation often requires numerical fidelity, domain-specific notation, spatial and relational reasoning, cross-panel evidence integration, and experimental context. Models that perform well on general captioning or visual question answering (VQA) may still misread axes and legends, overlook visual relations, or produce conclusions unsupported by the figure. Materials-science chart-to-table studies make this gap especially clear: advanced multimodal models may identify axes, legends, and overall trends while hallucinating or omitting data points in dense plots with overlapping markers or fitted curves \cite{circiinformation}. Reliable scientific-image understanding therefore requires both semantic interpretation and numerically faithful evidence extraction. These limitations constrain applications such as structured experimental-data extraction, evidence-grounded search, visual literature synthesis, and scientific question answering. Although systems such as ORKGEx already incorporate vision--language methods into scholarly knowledge-curation workflows \cite{hussein2025orkgex}, domain-grounded benchmarks remain necessary for evaluating their reliability.

The ALD/E-ImageMiner benchmark and the \href{https://sites.google.com/view/sci-imageminer/}{ICDAR 2026 Competition on Information Extraction from Atomic Layer Deposition/Etching Scientific Figures} provide a concrete response to this need \cite{ahmed2026icdar}. The benchmark brings together expert-annotated figures from experimental and simulation-based atomic layer deposition and etching (ALD/E) literature and evaluates four complementary capabilities: figure classification, data table extraction, summarization, and VQA. The competition report describes the dataset construction, annotation process, evaluation protocols, participating systems, and results. In the present vision paper, we use the benchmark and the experience gained from organizing the competition as a foundation for considering what future scientific-image challenges should measure and how their scope could develop.

Within these companion proceedings, this contribution complements the competition report \cite{ahmed2026icdar} and the participating-team studies by connecting the current challenge to a longer-term research agenda. We first examine ALD/E-ImageMiner as a testbed for capabilities ranging from visual localization and quantitative reading to domain-grounded interpretation, reasoning, and evidential justification. We then discuss how Bloom-informed question design can support different forms of engagement with the same scientific evidence and introduce \emph{scientific conceptual understanding from images} as a broader benchmark objective. Finally, we outline future extensions across materials science and related engineering domains, richer visual representations, contextual and cross-document synthesis, hypothesis evaluation, provenance, uncertainty, counterfactual grounding, and open-ended multimodal research.

\section{Related Work: Gaps in Scientific Image Understanding}
\label{sec:related-work}

\subsection{Benchmarks for Chart and Scientific Figure Understanding}

Chart-understanding benchmarks have established important tasks for extracting and reasoning over visually represented data, but much of this work has focused on synthetic or general-purpose visualizations. FigureQA \cite{kahou2017figureqa} and DVQA \cite{kafle2018dvqa} use programmatically generated charts paired with template-based questions, while PlotQA \cite{methani2020plotqa} provides a large collection of automatically generated plots and associated question-answer pairs. StructChart extends this line of work through SimChart9K, a language-model-driven collection of synthetic chart--table pairs intended to support structured chart understanding \cite{xia2023structchart}. ChartQA \cite{masry2022chartqa} introduced 20,882 real-world charts with human-authored questions, thereby broadening the visual and linguistic diversity of chart question answering; however, its web-sourced economic, social, and survey visualizations differ substantially from the specialized plots, diagrams, spectra, and composite figures found in scientific publications.

More directly related to materials science, Circi et al. introduce PolyCompChartIE and MetalThermoChartIE, two human-annotated benchmarks for chart-to-table extraction from real scatter plots concerning polymer composites and the thermophysical properties of metal systems \cite{circiinformation}. Their figures contain heterogeneous axis ranges, dense or overlapping data points, fitted curves, and categorical information represented through marker shapes, colors, and legends. The work also proposes Relative Coordinate-Label Similarity (RCLS), which evaluates whether extracted tables preserve the correspondence among numerical coordinates, axis labels, and series labels. This demonstrates that domain-specific chart understanding requires both representative scientific data and evaluation measures aligned with the structure of the visual evidence.

Together, these benchmarks have advanced chart classification, chart-to-table conversion, and visual question answering, while also revealing the additional demands introduced by scientific figures. Their meaning may depend on domain-specific notation, experimental conditions, measurement techniques, spatial organization, and relationships among several panels or modalities. A spectrum, microscopy image, reaction scheme, process diagram, or apparatus schematic cannot always be interpreted through the representations and question templates used for conventional statistical charts. Scientific-image benchmarks must therefore combine visual diversity with domain knowledge, numerically appropriate evaluation, and task definitions that reflect how evidence is communicated and used within scientific practice.

\subsection{Limitations of Vision--Language Models on Scientific Visual Evidence}

Recent evaluations indicate that strong performance on general visual benchmarks does not reliably transfer to scientific figures. MaCBench \cite{alampara2025probing}, which evaluates vision--language models on chemistry and materials-science problems, reports strong performance on tasks such as laboratory-equipment recognition and direct numerical extraction, but substantially weaker results when spatial reasoning, cross-modal synthesis, or multi-step inference is required. The MAC benchmark \cite{jiang2025mac}, constructed from scientific journal covers and their associated stories, similarly shows that models can recognize salient entities while relying heavily on textual cues; their performance declines when those cues are removed and interpretation must depend more directly on the visual content. More general investigations of the visual binding problem also show that models such as GPT-4V can fail to reliably associate objects, attributes, and spatial relations even in comparatively simple scenes \cite{campbell2024understanding}. Together, these findings suggest that visual recognition and textual fluency do not by themselves ensure reliable interpretation of scientific evidence.

Evaluations focused on particular visual forms reinforce this conclusion. Comparisons of GPT-4V and Gemini identify difficulties with table reasoning and formula recognition despite strong general captioning and question answering \cite{qi2023gemini}. ChartInsights \cite{wu2024chartinsights} reports considerable weaknesses in fine-grained chart analysis, while FlowLearn \cite{pan2024flowlearn} shows that model performance varies substantially across flowchart tasks such as OCR, node identification, and structural interpretation. Evaluations on real materials-science scatter plots further show that models may recover axes and legend information while producing incomplete or numerically inaccurate tables, with performance deteriorating on denser and more visually complex figures \cite{circiinformation}. Domain adaptation can improve results: visual instruction tuning supports stronger general multimodal performance \cite{liu2023visual}, and LLaVA-Med \cite{li2023llavamed} demonstrates the value of training on biomedical figures. Nevertheless, domain-adapted systems may still hallucinate visual details or fail on questions requiring several inferential steps. These results motivate benchmarks that evaluate complementary capabilities on the same domain-specific figures rather than relying on a single aggregate measure of visual understanding.

\subsection{Multimodal AI for Scholarly Knowledge Infrastructures}

Scientific-image understanding is also becoming relevant to digital-library systems that seek to transform publications into structured and searchable knowledge. ORKGEx \cite{hussein2025orkgex}, for example, combines language and vision models with the Open Research Knowledge Graph \cite{auer2020improving} to support the annotation of scholarly contributions. Its workflow recognizes that conventional OCR is insufficient for figures in which text, graphical marks, spatial relations, and domain-specific symbols jointly express meaning, and it incorporates vision--language methods for tasks including figure classification, chart-to-table conversion, and summarization. This illustrates how multimodal models can contribute to scholarly knowledge curation rather than serving only as standalone VQA systems.

However, the integration of scientific visual evidence into digital infrastructures remains at an early stage. Existing systems generally provide limited support for domain-specific interpretation, panel-level grounding, numerically faithful extraction, and the joint evaluation of several capabilities on the same figures. The literature therefore reveals a gap between general chart benchmarks, evaluations that expose weaknesses in scientific visual reasoning, and digital-library tools intended to produce structured scholarly knowledge. ALD/E-ImageMiner addresses this intersection through an expert-annotated, domain-grounded benchmark covering heterogeneous scientific figures and four complementary tasks: figure classification, data table extraction, summarization, and visual question answering \cite{ahmed2026icdar}. Next we examine how these tasks provide a testbed for scientific visual intelligence and a foundation for future challenge design.

\section{ALD/E-ImageMiner as a Testbed for Scientific Visual Intelligence}
\label{sec:sci-imageminer-testbed}

ALD/E-ImageMiner is not merely a collection of scientific figures, but a testbed for evaluating increasingly demanding vision-language capabilities in scientific settings. It contains 1,951 figures from 205 experimental and simulation-based ALD/E publications, organized into 49 figure categories \cite{ahmed2026icdar}. These include line and multi-line charts, spectra, scatter plots, heatmaps, molecular-structure and reaction diagrams, band and apparatus diagrams, process flows, and composite image panels. The Hugging Face release enables systematic browsing by figure type: \url{https://huggingface.co/datasets/SciKnowOrg/ALD-E-ImageMiner}. This diversity reflects the many ways scientific evidence is encoded, including through axes and legends, spectral peaks, chemical notation, spatial relationships, and domain-specific visual conventions. Further details on the benchmark, annotations, evaluation, competition, systems, and results are provided in the ICDAR competition report \cite{ahmed2026icdar}. In this paper, we use the benchmark as a basis for articulating desiderata for future vision-language model research, including the design of scientific benchmarks.

\subsection{From Seeing to Scientific Reasoning}

Scientific figure comprehension can be understood as a progression from \emph{seeing} to \emph{reading}, \emph{understanding}, \emph{reasoning}, and ultimately \emph{justifying}. First, a model must see the relevant visual structure by distinguishing subfigures, axes, legends, labels, symbols, curves, regions, and other graphical elements. It must then read the figure with sufficient quantitative fidelity to recover values, units, categories, and relations without losing their structural organization. To understand the figure, the model must connect these observations to the represented scientific variables, experimental conditions, processes, and domain terminology. Scientific reasoning further requires deriving relationships and implications that cannot be obtained through direct transcription alone, including causal mechanisms, comparative trends, structure--property relations, and application-oriented conclusions. A trustworthy system should additionally be able to justify its response by locating the relevant visual evidence, distinguishing direct observations from domain-informed interpretations, and expressing uncertainty when the available evidence is insufficient.

The four ALD/E-ImageMiner tasks provide complementary operationalizations of this progression. \textsc{task 1.} \textit{Figure classification} probes whether a model can identify the representational form through which information is communicated. \textsc{task 2.} \textit{Data table extraction} tests whether quantitative values and their relations can be recovered with structural and numerical fidelity. \textsc{task 3.} \textit{Summarization} examines whether a model can move beyond transcription to identify the central scientific message conveyed by a figure. \textsc{task 4.} \textit{Visual question answering} (VQA) then targets the most explicit transition from visual perception to domain-grounded scientific reasoning.

\subsection{Bloom-Informed Visual Question Answering}

The VQA component was deliberately designed using Bloom's revised taxonomy as a scaffold for eliciting progressively deeper levels of reasoning \cite{krathwohl2002revision}. The questions are grounded in the scientific characteristics of atomic layer deposition and etching (ALD/E), where material growth or removal is controlled through cyclic, surface-mediated processes. Figures in this literature therefore communicate more than isolated numerical values: they describe precursor and co-reactant exposures, purge and reaction steps, process windows, growth-per-cycle behavior, changes in composition and structure, and the resulting material or device performance \cite{ahmed2026icdar}. Understanding such figures may require a model to trace a sequence of surface reactions, relate processing conditions to measured trends, connect film structure or composition to physical properties, or interpret the implications of those properties for a particular application. 

In contrast to generic VQA datasets dominated by object recognition or direct lookup, the questions in ALD/E-ImageMiner are organized around four domain-grounded families: \emph{process-oriented}, \emph{comparative/trend}, \emph{structure--property}, and \emph{application/performance} questions. Together, these families span cognitive operations ranging from remembering and understanding visible information to applying domain concepts, analyzing relationships, and evaluating scientific or technological implications.

Bloom's taxonomy is used here as a question-design framework without implying that model computation is equivalent to human cognition. Moreover, the question families do not correspond rigidly to individual Bloom levels. The same figure can support a direct factoid question, an analytical comparison, a causal explanation, or an evaluative judgment. This makes it possible to vary not only the visual content presented to a model, but also the depth at which that content must be interpreted.

\begin{description}

    \item[\textbf{Process-oriented questions.}] These questions concern the cyclic and surface-mediated nature of ALD/E, including precursor exposures, purge steps, surface reactions, modification and removal stages, and complete deposition or etching workflows. A sequencing question may primarily involve \emph{Remembering} and \emph{Understanding}; explaining why a purge step is necessary requires \emph{Understanding} and \emph{Analyzing}; and predicting the consequence of changing a pulse or purge condition requires \emph{Applying} scientific knowledge to a new situation.

    \item[\textbf{Comparative/trend questions.}] These questions compare experimental variables and determine how changes in temperature, pulse length, cycle count, precursor ratio, or composition affect outcomes such as growth per cycle, etch per cycle, film thickness, or spectral intensity. Directly reporting a trend may involve \emph{Understanding}, whereas explaining correlations, identifying competing effects, or predicting behavior under an unobserved condition requires \emph{Applying} and \emph{Analyzing}. Comparisons among alternative process routes can additionally elicit \emph{Evaluation}.

    \item[\textbf{Structure--property questions.}] These questions ask models to connect precursor chemistry, film composition, molecular or layer structure, doping arrangement, or interface organization with resulting material properties. They therefore move beyond identifying visible entities and require the model to \emph{Apply} domain knowledge and \emph{Analyze} how changes in structure or composition influence electronic, optical, chemical,  mechanical, or surface behavior.

    \item[\textbf{Application/performance questions.}] These questions relate experimentally observed material behavior to device-level performance or practical use. They may require a model to determine whether a spectral change indicates improved emission, whether a chromaticity coordinate is relevant to a lighting application, or how a measured material response could support sensing, energy conversion, or semiconductor-device fabrication. Such questions primarily target \emph{Analysis} and \emph{Evaluation}, because the answer must connect visual evidence to a broader scientific or technological objective.

\end{description}

Questions are paired with four answer formats: yes/no, factoid, list, and paragraph. These formats permit different levels of granularity while avoiding the assumption that all scientific reasoning must be expressed as long-form text. Factoid and list answers can test precise recognition, quantitative recovery, or process sequencing, whereas paragraph answers require models to make their reasoning and interpretation more explicit. The annotation design includes explanatory paragraph answers so that the benchmark can evaluate not only whether a conclusion is correct, but also whether it is scientifically coherent and grounded in the figure.

The following multi-panel example illustrates how this Bloom-informed VQA design is instantiated through panel-level annotation and machine-readable visual grounding in ALD/E-ImageMiner.

\subsection{A Multi-Panel Example from ALD/E-ImageMiner}
\label{sec:multitask-example}

\begin{figure}[!htb]
    \centering
    \includegraphics[width=.7\linewidth]
    {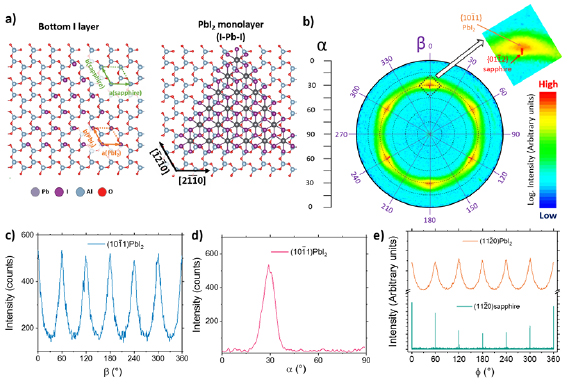}
    \caption{Reproduced from Figure~3 in Mattinen et al., \emph{Van der Waals epitaxy of continuous thin films of 2D materials using atomic layer deposition in low temperature and low vacuum conditions} \cite{mattinen2020van}. The ALD/E-ImageMiner classification annotations identify (a)~a molecular structure diagram, (b)~a polar chart (rose chart), (c--d)~spectra charts, and (e)~a stacked spectra chart. The figure also supports domain-grounded questions concerning epitaxial growth, crystallographic orientation, and disorder in PbI$_2$ films grown on sapphire.}
    \label{fig:ald-e-multipanel-example}
\end{figure}

\autoref{fig:ald-e-multipanel-example} provides a concrete example of how the benchmark combines heterogeneous scientific visual content with panel-specific task annotations. During dataset preparation, \href{https://github.com/opendatalab/MinerU}{MinerU} \cite{wang2024mineru} was first applied to extract the structured textual content and high-resolution figures from each publication. The document text and figure captions were stored in a top-level \texttt{content.json} file, while the extracted figures and their corresponding annotation files were organized separately. To facilitate machine reading of composite figures, each subfigure was manually associated with its panel letter and localized using machine-readable bounding-box coordinates \((x,y,\mathrm{width},\mathrm{height})\). This enabled annotations associated with a particular panel to be paired with the corresponding image region when presented to a vision-language model, rather than requiring the model to infer the target region from the complete multi-panel figure.

For this example, the five panels were localized using the coordinates shown in \autoref{tab:multipanel-bboxes}, expressed in pixels relative to the complete figure.

\begin{table}[!htb]
    \centering
    \small
    \caption{Panel-level bounding boxes for \autoref{fig:ald-e-multipanel-example}.}
    \label{tab:multipanel-bboxes}
    \begin{tabular}{crrrr}
        \hline
        \textbf{Panel} & \(\boldsymbol{x}\) & \(\boldsymbol{y}\) &
        \textbf{Width} & \textbf{Height} \\
        \hline
        (a) & 15  & 55  & 597 & 392 \\
        (b) & 623 & 40  & 508 & 436 \\
        (c) & 17  & 493 & 374 & 286 \\
        (d) & 393 & 488 & 363 & 290 \\
        (e) & 778 & 492 & 360 & 286 \\
        \hline
    \end{tabular}
\end{table}

The panel identifiers link each task annotation to its intended subfigure, while the coordinates allow the corresponding visual region to be deterministically isolated. For example, a question associated with panel~(b) can be supplied together with the crop defined by \((623,40,508,436)\). The bounding boxes are not assumed to improve model performance by themselves; rather, they provide explicit machine-readable grounding so that the annotation, visual input, and evaluation consistently refer to the intended panel.

This figure was annotated with four domain-grounded questions covering three question families and multiple answer formats:

\begin{description}

    \item[\textbf{Panel (a): Process-Oriented; Yes/No.}] 
    \emph{Question:} Can PbI$_2$ be grown epitaxially over sapphire?

    \emph{Answer:} Yes.

    \item[\textbf{Panel (b): Structure--Property; Paragraph.}]
    \emph{Question:} Why was the pole figure measured at
    \(25.95^{\circ}\)?

    \emph{Answer:} The pole figure measured at \(25.95^{\circ}\,2\theta\)
    corresponds to the asymmetric
    \((1\,0\,\bar{1}\,1)\) PbI$_2$ reflection, but is also sufficiently close
    to the asymmetric \((0\,1\,\bar{1}\,2)\) sapphire reflection to provide
    information about the film--substrate registration. Three sharp peaks over
    a \(360^{\circ}\) rotation were observed for the sapphire reflection,
    whereas six broader peaks were observed for PbI$_2$; considering the
    symmetry of the reflections, both patterns are consistent with a single
    in-plane orientation. The alignment of the
    \((1\,0\,\bar{1}\,1)\) PbI$_2$ and
    \((0\,1\,\bar{1}\,2)\) sapphire reflections corresponds to the epitaxial
    relation
    \[
        (0\,0\,1)[2\,\bar{1}\,1\,0]_{\mathrm{PbI_2}}
        \parallel
        (0\,0\,1)[2\,\bar{1}\,1\,0]_{\mathrm{Al_2O_3}},
    \]
    indicating parallel alignment of the unit-cell axes of PbI$_2$ and
    sapphire, as expected from the relatively small lattice mismatch between
    the film and substrate.

    \item[\textbf{Panel (d): Comparative/Trend; Factoid.}]
    \emph{Question:} What suggests disorder in the PbI$_2$ films?

    \emph{Answer:} The relatively broad peaks in the
    \(\beta\) scan (\(\mathrm{FWHM}=13^{\circ}\), panel~(c)) and the
    \(\alpha\) scan (\(\mathrm{FWHM}=9^{\circ}\), panel~(d)), extracted from
    the in-plane pole figure, suggest disorder in the in-plane and
    out-of-plane orientations, respectively.

    \item[\textbf{Panel (e): Structure--Property; Factoid.}]
    \emph{Question:} Are there \(30^{\circ}/90^{\circ}\) domains in
    PbI$_2$ grown on sapphire?

    \emph{Answer:} In-plane XRD measurements confirmed the in-plane
    alignment, the absence of \(30^{\circ}/90^{\circ}\) domains, and the
    presence of some non-epitaxial domains.

\end{description}

This example illustrates how the benchmark combines panel-level visual grounding with different forms of scientific interpretation. A model may need to recognize the representational form of an individual panel, relate measurements across panels, and interpret the evidence using concepts such as epitaxy, film--substrate registration, crystallographic orientation, and structural disorder. The complete annotation record, including the summarization and data-table extraction targets, is available at \url{https://github.com/sciknoworg/ALD-E-ImageMiner/blob/main/icdar2026-competition-data/train/atomic-layer-deposition/experimental-usecase/37/images/figure_3.json} \cite{ahmed2026icdar}.

\subsection{Toward Create-Level Scientific Visual Reasoning}

The preceding example shows how the current ALD/E-ImageMiner annotations progress from visual recognition to the analysis and evaluation of scientific evidence represented in a figure. A natural direction for future challenge editions is to extend this progression toward \emph{Create}, the highest level in Bloom's revised taxonomy, by asking models to formulate scientifically plausible experimental workflows or material designs under explicit constraints. Such tasks would require models to synthesize distributed visual information into an ordered and coherent plan while distinguishing supported steps from parameters not specified in the available evidence, such as pulse duration, deposition temperature, target thickness, or donor--acceptor distance. Their evaluation could consider the inclusion and ordering of essential stages, consistency with the visual and textual evidence, scientific feasibility, treatment of missing information, and avoidance of unsupported procedural details. The following hypothetical question, based on Figure~1 of Ghazy et al. \cite{ghazy2023atomic}, illustrates this prospective extension and is not part of the current ALD/E-ImageMiner annotations.

\begin{quote}
\small
\textbf{Question:} Based on the illustrated process and optical configuration, outline a minimal experimental workflow for preparing and testing a Eu--HQA hybrid layer for a FRET-based application.

\textbf{Answer:}
\begin{enumerate}
    \item Repeat the following ALD/MLD cycle until the target thickness is reached: Eu(thd)\(_3\) pulse \(\rightarrow\) N\(_2\) purge \(\rightarrow\) HQA pulse \(\rightarrow\) N\(_2\) purge.

    \item Deposit the hybrid layer either on a plasmonic nanostructure for emission enhancement or on flat Si as a reference.

    \item Introduce AF647 under conditions that control the donor--acceptor spacing required for measurable FRET.
\end{enumerate}
\end{quote}

More broadly, future challenge editions could introduce paired questions of increasing cognitive depth for the same figure, complemented by cross-panel reasoning, provenance-aware answers, and uncertainty assessment. These additions would enable a richer evaluation of scientific visual reasoning.

\section{Toward ``Scientific Conceptual Understanding'' from Images: A Challenge Roadmap}
\label{sec:conceptual-understanding}

We propose \emph{scientific conceptual understanding from images} as a long-term objective for multimodal AI and digital libraries. This capability extends beyond recognizing visual forms, extracting values, or producing plausible descriptions. It requires a system to use visual evidence within a scientific process: to identify what was observed, relate it to experimental conditions and domain knowledge, compare it with alternative evidence, and determine what conclusions are warranted. The current ALD/E-ImageMiner tasks provide complementary and independently evaluable probes of this objective. Classification identifies how evidence is represented, data extraction recovers machine-readable quantitative content, summarization captures the principal message of a figure, and VQA evaluates domain-grounded interpretation and reasoning.

Building on this foundation, we outline two directions for the future development of Sci-ImageMiner. The first concerns the incremental expansion of its figure types and scientific domains, initially across materials science and subsequently into related engineering disciplines. The second concerns new task objectives that connect scientific images with contextual interpretation, cross-figure synthesis, hypothesis evaluation, provenance, uncertainty, and broader research workflows.

\subsection{Figure-Type and Domain Extensions}
\label{sec:figure-domain-extensions}

The first direction is to expand ALD/E-ImageMiner incrementally while preserving its expert-annotated and domain-grounded character. Within materials science, the nearest extensions include thin-film synthesis and characterization, semiconductor processing, two-dimensional materials, catalysis, electrochemistry and battery materials, photovoltaics, polymers, ceramics, and composite materials. These areas share many of the experimental techniques and processing--structure--property relations already encountered in ALD/E, while introducing different materials, mechanisms, and performance criteria. Such a progression would make it possible to evaluate whether a model transfers its understanding of familiar visual conventions to related scientific settings. It would also support concrete applications such as retrieving experiments conducted under comparable conditions, comparing fabrication routes across material classes, constructing structured materials-property records, and identifying earlier studies that report similar structural or performance trends.

The expansion should be guided by the scientific functions of visual representations rather than by visual diversity alone. Microscopy images from scanning electron microscopy (SEM), transmission electron microscopy (TEM), and atomic force microscopy (AFM), and related techniques could support the characterization of morphology, interfaces, defects, particle distributions, and film coverage. Diffraction patterns, reciprocal-space maps, and spectroscopic figures could support phase identification, crystallographic-orientation analysis, peak assignment, and comparison of chemical or structural states. Phase diagrams and multidimensional property maps could be converted into searchable process windows, stability regions, and composition--property relations, while atomistic models, band structures, density-of-states plots, device cross-sections, reaction schemes, process diagrams, and apparatus schematics could represent mechanisms, configurations, and experimental workflows that cannot be adequately reduced to ordinary tables. The resulting benchmark would test whether models can transform different visual languages into scientifically useful representations while preserving  quantities, relations, constraints, and uncertainty. Related work on chemical structure recognition provides a concrete example: molecular and Markush structure images are translated into machine-readable SMILES or CXSMILES records encoding atoms, bonds, variable sites, and substituent constraints \cite{andonian2026markushglyph}. Such annotations would extend the benchmark toward chemical diagrams while enabling programmatic structure search, database indexing, and the construction of training data for downstream chemistry models.

After deepening its materials-science coverage, the dataset could expand into related engineering sciences that communicate knowledge through similarly specialized visual forms. Chemical and process engineering would introduce reactor schematics, process flowsheets, transport plots, and kinetic models; electrical and electronic engineering would contribute circuits, device architectures, semiconductor cross-sections, and timing diagrams; mechanical and aerospace engineering would add stress fields, fracture images, simulation outputs, and CAD-like representations; and energy or biomedical engineering would provide system configurations, diagnostic images, and device-performance figures. A modular annotation framework could preserve a shared core of panel localization, classification, structured extraction, summarization, VQA, and source provenance while adding schemas and question families specific to each field. Metadata describing the domain, subdomain, visual form, experimental or computational setting, and required background knowledge would permit controlled comparisons between in-domain performance, transfer to neighboring fields, and generalization across substantially different scientific conventions.

\subsection{Task-Objective Extensions}
\label{sec:task-objective-extensions}

The scientific research lifecycle spans literature discovery, evidence synthesis, hypothesis formulation, experimentation, content production, and evaluation \cite{eger2025transforming}. Future Sci-ImageMiner challenges could connect figure understanding to these wider research activities while retaining clear task-wise evaluation and extending the benchmark toward increasingly contextual, comparative, and open-ended uses of scientific visual evidence.

\paragraph{Context-grounded and machine-actionable interpretation.}
A contextual figure-understanding task could provide a figure together with selected captions, methods, and surrounding passages and require the model to distinguish among information directly visible in the figure, information reported in the text, and conclusions requiring domain knowledge. The output could include both a natural-language interpretation and a structured record of variables, values, units, conditions, and relations. For example, a system might identify a spectral shift from the image, retrieve the corresponding deposition condition from the methods, and mark a proposed reaction mechanism as an interpretation rather than an observation. Such outputs would support figure-aware paper chat, scientific database population, evidence-grounded search, and the integration of visual observations into knowledge graphs.

\paragraph{Hypothesis formation, discriminating tests, and revision.}
A more demanding track could present visual evidence and ask a system to generate competing explanations, identify which additional measurement would best distinguish them, and revise its assessment when new evidence is introduced. For example, broad diffraction peaks might be explained by small crystallite size, strain, or orientational disorder; a model could be asked to identify which microscopy, diffraction, or complementary spectroscopy result would discriminate among these possibilities. The benchmark could then add a new panel and test whether the system updates its conclusion rather than rationalizing its initial answer. This would support experiment planning, measurement selection, materials troubleshooting, and the design of follow-up studies.

Such evaluation could use a structured \emph{epistemic record} rather than an unrestricted chain of thought. A record might contain the selected visual evidence, candidate hypotheses, predicted observations, chosen test, resulting evidence, and revised conclusion. These fields expose claims that can be checked against the figure and domain annotations without assuming that the record reproduces the model's internal computation. Bloom-informed question families could define the intended cognitive demand, while explicit evidence--hypothesis--test--revision links would measure whether the observable response follows a scientifically disciplined structure.

\paragraph{Cross-panel and cross-document synthesis.}
Many scientific conclusions depend on several complementary measurements. Cross-panel tasks could require models to connect a structural image with a spectrum, relate a control sample to a treated sample, or explain how changes in process conditions correspond to changes in measured properties. At the document level, a system could trace a material from synthesis and characterization to device evaluation. Cross-document tasks could then compare related figures from multiple publications, normalize experimental variables, and determine whether reported results agree, differ because of their conditions, or remain genuinely contradictory. These capabilities would support visual literature reviews, evidence synthesis, reproducibility analysis, and the construction of searchable processing--structure--property networks.

\paragraph{Counterfactual grounding, provenance, and uncertainty.}
Because plausible explanations may be unfaithful \cite{turpin2023language} and relevant computation need not be fully visible in generated text \cite{baherwani2026not}, explanation quality alone should not be used as evidence that a model reasoned correctly. Future tasks could instead introduce controlled counterfactuals: remove a relevant panel, alter a plotted value, exchange two labels, add contradictory evidence, or modify irrelevant formatting. A grounded model should change its answer when the scientific evidence changes, remain stable under irrelevant visual perturbations, and abstain when the necessary evidence is absent. Requiring each answer to identify its supporting panel, bounding box, plotted value, or document passage would further enable claim verification, figure-aware peer review, and transparent scholarly retrieval.

\paragraph{Open-ended multimodal research challenges.}
The eventual objective should extend beyond answering predetermined questions toward evaluating whether multimodal agents can use visual literature in an open-ended research process. Recent case studies suggest that agents can complete substantial research engineering while still struggling with experimental judgment, evidence selection, backtracking, and determining whether a research question has actually been resolved \cite{rios2026ai,kirgis2026can}. A future ``shadow'' challenge could give an agent a research question derived from an unpublished or carefully held-out study, together with a literature collection containing figures, tables, methods, and data. The agent could be asked to synthesize prior evidence, formulate hypotheses, select analyses or experiments, and produce a justified research proposal or result for assessment by domain experts. To preserve its multimodal focus, the task would center on reasoning over figures and tables, with textual passages, and associated data serving as supporting context.

Such open-ended evaluations should complement rather than replace automatically scored tasks. Classification, extraction, summarization, and VQA provide reproducible diagnostics of specific capabilities, while structured epistemic tasks evaluate evidence use and revision, and expert-assessed challenges probe scientific judgment under less constrained conditions. Together, these evaluation layers could distinguish a system that merely produces a plausible final answer from one that reliably connects visual observations, scientific claims, tests, and decisions.

The resulting vision is a continuously expanding multimodal scientific environment rather than a static collection of images. A future Sci-ImageMiner system could retrieve related visual evidence, transform it into machine-actionable knowledge, compare findings across publications, propose competing explanations, request the next informative measurement, revise its conclusions, and report what remains uncertain. Its progress would remain measurable through independently scored tasks, but its broader purpose would be to evaluate whether multimodal AI can participate responsibly in the evidence-driven and self-correcting practices of science.

\section{Conclusion}

Scientific figures and tables are primary carriers of experimental evidence, yet much of their content remains difficult to search, compare, and integrate within digital knowledge infrastructures. ALD/E-ImageMiner provides a domain-grounded testbed for addressing this gap through complementary tasks covering classification, structured data extraction, summarization, and visual question answering. Building on this foundation, we have proposed \emph{scientific conceptual understanding from images} as a broader benchmark objective: the ability to interpret visual evidence in context, relate it to scientific claims and experimental conditions, and determine which conclusions are supported. Rather than treating this objective as a single measure of ``understanding,'' the proposed roadmap decomposes it into independently evaluable capabilities that can be extended across figure types, scientific domains, and research activities.

The future development of Sci-ImageMiner should therefore proceed incrementally, first broadening its coverage within materials science and related engineering disciplines, and then introducing tasks involving contextual interpretation, cross-figure synthesis, hypothesis evaluation, provenance, uncertainty, and open-ended multimodal research. Such benchmarks could enable digital libraries to move beyond caption-based indexing toward machine-actionable representations of visual evidence, evidence-grounded retrieval, visual literature synthesis, and support for experimental planning and scientific evaluation. Achieving this vision will require sustained domain-expert annotation, transparent task design, and evaluation methods that reward numerical fidelity, evidential grounding, appropriate revision, and abstention when the available information is insufficient. In this way, scientific image understanding can become a concrete and progressively measurable foundation for general-purpose scientific AI that learns from the full multimodal scholarly record.

\section*{Data availability statement}

The ALD/E-ImageMiner benchmark data discussed in this contribution are publicly available at \url{https://github.com/sciknoworg/ALD-E-ImageMiner/tree/main/icdar2026-competition-data} and are described in the ICDAR 2026 competition report \cite{ahmed2026icdar}. The release includes scientific figure images, machine-readable annotations and metadata, structured content files, evaluation scripts, and submission guidelines. Source article PDFs are intentionally excluded from the GitHub distribution.

ALD/E-ImageMiner is released as a mixed-rights, non-commercial benchmark resource. The annotations and generated metadata are available under CC BY 4.0, whereas extracted figure images and source-derived content remain subject to the rights and reuse terms of their respective source articles. There is therefore no blanket open license covering the complete corpus. Users must consult the corresponding source publication and rights-holder terms before redistributing, reusing, or commercially exploiting individual images. Further details are provided in the repository-level license notice: \url{https://github.com/sciknoworg/ALD-E-ImageMiner/blob/main/LICENSE}.

\section*{Underlying and related material}

The ALD/E-ImageMiner competition project page provides access to the benchmark dataset, documentation, and related resources:

\begin{quote}
\url{https://sciknoworg.github.io/ALD-E-ImageMiner/}
\end{quote}

Codabench submission-format guidelines and the official evaluation scripts used on the Codabench competition pages are available at:

\begin{itemize}
    \item Submission-format guidelines:
    \url{https://github.com/sciknoworg/ALD-E-ImageMiner/tree/main/icdar2026-competition-data/test/submission_guidelines}

    \item Official competition evaluation scripts:
    \url{https://github.com/sciknoworg/ALD-E-ImageMiner/tree/main/icdar2026-competition-data/evaluation_scripts}
\end{itemize}

The live Codabench leaderboards remain open indefinitely for submissions, subject to the continued availability of the Codabench platform. The competition pages for the respective tasks are listed below.

\begin{itemize}
    \item Classification task:
    \url{https://www.codabench.org/competitions/12901/}

    \item Data table extraction task:
    \url{https://www.codabench.org/competitions/12902/}

    \item Summarization task:
    \url{https://www.codabench.org/competitions/12909/}

    \item Visual question answering task:
    \url{https://www.codabench.org/competitions/12908/}
\end{itemize}

\section*{Author contributions}

J.D.: Conceptualization, Funding acquisition, Methodology, Project administration, Resources, Supervision, Writing -- original draft. F.A.: Conceptualization, Methodology, Project administration, Resources, Software. C.A.B.A., L.F., P.G., D.H., M.U.K., N.K.N., P.P.J., F.P., S.F.v.d.W., and T.F.J.v.R.: Data curation, Investigation, Validation. 

Authors 3--12 contributed as domain-expert dataset annotators and are listed alphabetically by family name.

\section*{Competing interests}

The authors declare that they have no competing interests.

\section*{Funding}

The creation of the ALD/E-ImageMiner benchmark dataset was supported by the
\href{https://www.nfdi4datascience.de/}{NFDI4\allowbreak DataScience}
initiative, funded by the German Research Foundation
(DFG, Grant ID: 460234259).

\section*{Acknowledgements}
We would like to thank all participants in the Sci-ImageMiner competition for their valuable contributions, thoughtful engagement, and the scientific exchange that made the competition a particularly enriching experience. We also thank the \href{[https://icdar2026.org/index.php/competitions/}{ICDAR 2026 competition chairs} for the opportunity to host the competition as part of ICDAR 2026.
We also thank Eleni Poupaki (TUE, NL), Alex Watkins (UOW, UK), Bora Karasulu (UOW, UK), Adrie Mackus (TUE, NL), Erwin Kessels (TUE, NL), and Vijay K. Narasimhan (M Ventures, USA) for extended discussions. These scientists, together with the first author, participated in the ``AI-Aware Pathways to Sustainable Semiconductor Process and Manufacturing Technologies (AWASES)'' program, funded by Merck and Intel.

\printbibliography[heading=references]

\end{document}